\documentclass[10pt,twocolumn,letterpaper]{article}

\usepackage[ruled]{algorithm2e}
\usepackage{algorithmic}
\usepackage{cvpr}
\usepackage{times}
\usepackage{epsfig}
\usepackage{graphicx}
\usepackage{amsmath}
\usepackage{amssymb}
\usepackage{amsfonts}
\usepackage{amsbsy}
\usepackage{amsmath}
\usepackage{booktabs}
\usepackage[table]{xcolor}
\usepackage{array}
\usepackage{makecell}
\usepackage{cellspace}
\usepackage{mathtools}
\usepackage{authblk}
\usepackage{multirow}
\usepackage[pagebackref=true,breaklinks=true,letterpaper=true,colorlinks,bookmarks=false]{hyperref}

 \cvprfinalcopy % *** Uncomment this line for the final submission

\def\cvprPaperID{9573} % *** Enter the CVPR Paper ID here

\begin{document}

%%%%%%%%% TITLE
\title{Investigating Vulnerability to Adversarial Examples on Multimodal Data Fusion in Deep Learning\vspace{-5mm}}

\author{Youngjoon Yu, Hong Joo Lee, Byeong Cheon Kim, Jung Uk Kim, and Yong Man Ro\\\vspace{-0.2cm}
	Image and Video Systems Lab, School of Electrical Engineering, KAIST, South Korea\\
	{\tt\small \{greatday, dlghdwn008, bckim1318, jukim0701, ymro\}@kaist.ac.kr}
	% For a paper whose authors are all at the same institution,
% For a paper whose authors are all at the same institution,
% omit the following lines up until the closing ``}''.
% Additional authors and addresses can be added with ``\and'',
% just like the second author.
% To save space, use either the email address or home page, not both
}

\maketitle
\thispagestyle{empty}

%%%%%%%%% ABSTRACT
\begin{abstract}
The success of multimodal data fusion in deep learning appears to be attributed to the use of complementary information between multiple input data. Compared to their predictive performance, relatively less attention has been devoted to the robustness of multimodal fusion models. In this paper, we investigated whether the current multimodal fusion model utilizes the complementary intelligence to defend against adversarial attacks. We applied gradient based white-box attacks such as FGSM and PGD on MFNet, which is a major multispectral (RGB, Thermal) fusion deep learning model for semantic segmentation. We verified that the multimodal fusion model optimized for better prediction is still vulnerable to adversarial attack, even if only one of the sensors is attacked. Thus, it is hard to say that existing multimodal data fusion models are fully utilizing complementary relationships between multiple modalities in terms of adversarial robustness. We believe that our observations open a new horizon for adversarial attack research on multimodal data fusion. 
\end{abstract}

%%%%%%%%% BODY TEXT
\section{Introduction}
Deep learning models can have multiple types of data as inputs. When a number of data types are being engaged in the processing of information, it is called multimodal data fusion models \cite{kanezaki2019deep,gao2020survey}. Multimodal data fusion deep learning models have focused on improving the performance of the model. Efforts to utilize multimodal data to improve the prediction performance of deep learning have been made in a number of tasks. Many studies in computer vision took account of multimodal data fusion when researchers want to enhance the performance such as for medical image segmentation \cite{du2016overview,guo2019deep}, semantic segmentation \cite{guo2018semantic,ramachandram2017deep}, and person re-identification \cite{tao2017deep,pala2015multimodal}.

However, relatively less attention has been devoted to the research on the modality of the deep learning model in adversarial attack and robustness. We intend to fill this research gap between multimodal data fusion and the adversarial robustness. In previous studies, it has been known that the performance enhancement of multimodal data fusion deep learning is attributed to the complementary effects between multiple modality data \cite{hazirbas2016fusenet}. In our experiment, the fusion deep learning model using multimodality was as vulnerable to adversarial attacks as a single-modal model. Therefore, it is hard to say that existing multimodal deep learning models are fully utilizing complementary relationships between multiple modality data.

In this paper, we consider the problem of adversarial vulnerabilities in the multimodal context. Our experiments show that even if only one of the multimodal inputs is attacked, it is significantly lethal to the entire performance of the model. Recent studies in the adversarial attack research, the increase of input dimensions \cite{ford2019adversarial,simon2018adversarial} has a negative impact on adversarial robustness. Overall, the contributions of this study are mainly in following two aspects:
\begin{itemize}
    \item First, we analyze and spotlight the adversarial robustness of multimodal data fusion. We explore the possible adversarial vulnerabilities in the multimodal deep learning model. We examined how effective the complementary relationship of current multimodal data fusion models in the sense of adversarial robustness.
    \item Our paper also presents the empirical evidence of the adversarial vulnerability of multimodal data fusion. We conducted experiments to find how adversarial robustness changes with different attacks, different epsilon perturbations in a multimodal data fusion model. By combining the experimental results with theoretical explanations, we are able to find a better interpretation and improvement on the adversarial vulnerability of multimodal data fusion.
\end{itemize}

The rest of the paper is organized as follows. In section 2, related works are introduced. In section 3, experimental settings and results are presented. The conclusion is drawn in section 4.

\section{Related Works}
The multimodal data fusion has become a rapidly growing research field (image–word, image–audio, multiple image sensor data). In this paper, we focus on handling multiple image sensor data based multimodal data fusion method. Single-task and multitask learning methods with multimodal data focused on obtaining better shared representation from multimodal data. By doing so, the model receives complementary information from other modal and shows consistent performances when there is a lack of particular information \cite{ha2017mfnet,hazirbas2016fusenet,sun2019rtfnet}. Supplementing missing information allows the model to predict more accurate. However, there has been limited analysis on the adversarial attack and defense in the multimodal context. In this paper, we analyzed the vulnerability of multimodal data fusion deep learning models. 
\subsection{Multimodal Encoder-Decoder Networks}
Among many different tasks, we focused on a semantic segmentation so that we could evaluate the multimodal data fusion model in an adversarial robustness context. For efficient multitask learning with a shared feature representation, multimodal encoder-decoder network was introduced. \cite{badrinarayanan2017segnet} Since shared representations from different multiple modalities enhance the performance of models, multimodal encoder-decoder networks are commonly used for the semantic segmentation. \cite{badrinarayanan2017segnet,hazirbas2016fusenet,sun2019rtfnet}
\subsection{Adversarial Attack}
For semantic segmentation tasks, the multimodal approaches performed well in terms of prediction accuracy. However, they exposed to serious vulnerabilities by the adversarial attacks below. Adversarial attacks are perturbations that cause misclassification and malfunctions of deep learning network. Among a variety of adversarial attacks, two gradient based methods were used in our experiment, which were found to work effectively in the semantic segmentation task.

\textbf{Fast Gradient Sign Method (FGSM)} \cite{goodfellow2014explaining}. Goodfellow \etal (2014) introduces a one-step attacking method, which crafts an adversarial example $x'$ as 
\begin{equation}
x'=x+ \epsilon*sign(\bigtriangledown_{x} L(\theta ,x,y)),    
\end{equation}
with the perturbation $\epsilon$ and the training loss L($\theta$,x,y).

\begin{figure*}[t]
\label{fig:1}
	\begin{minipage}[b]{1.0\linewidth}
		\centering
		\centerline{\includegraphics[width=17.5cm]{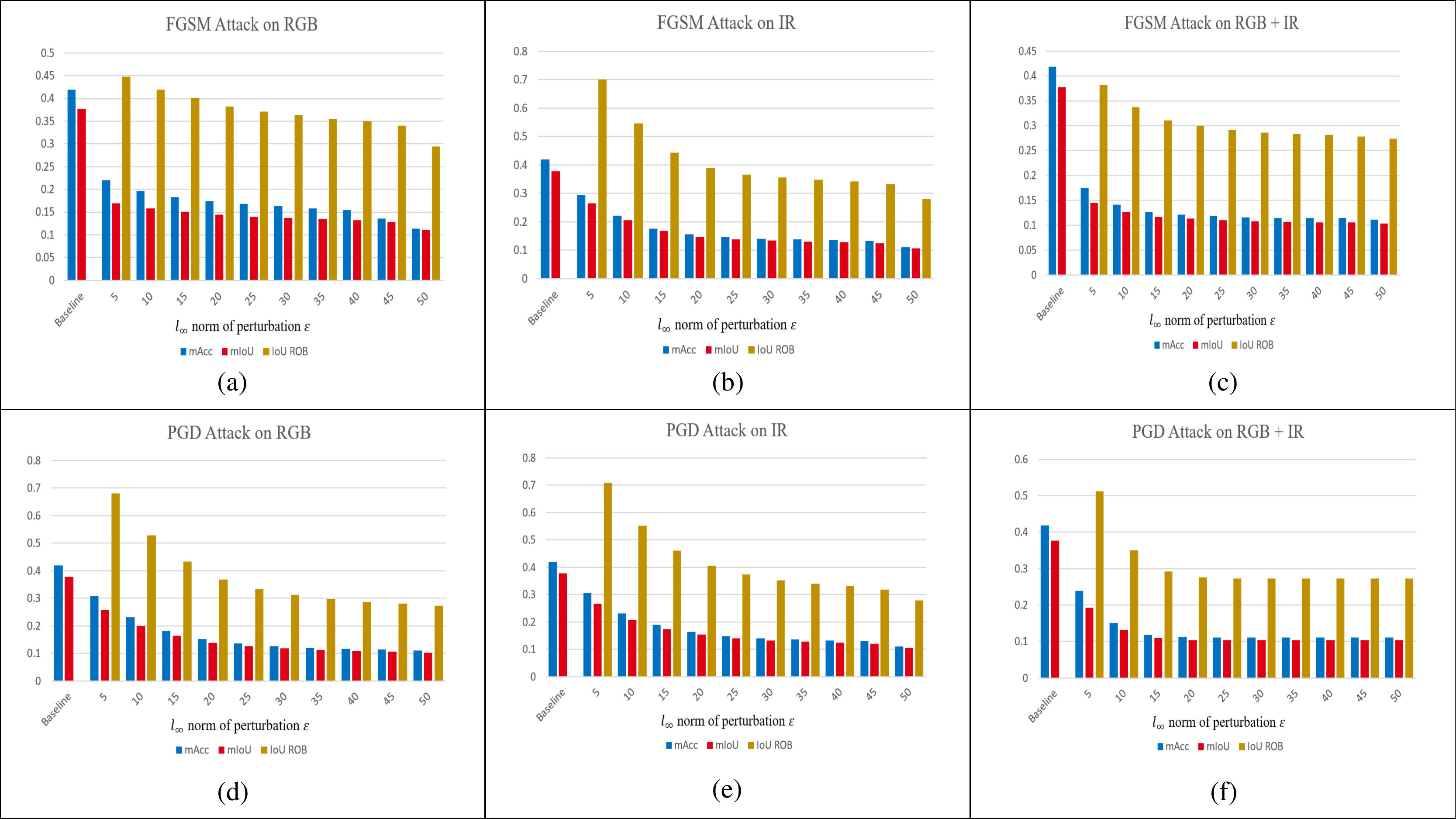}}
	\end{minipage}
	\vspace{-0.3cm}
	\caption{Evaluations of adversarial attacks on MFNet multimodal data fusion model. The $IoU_R$ value of the baseline model equals to one. (a), (b) and (c) denote the FGSM attack and (d), (e), and (f) refer to the PGD attack results.}
	\vspace{-0.1cm}
\end{figure*}

\textbf{Projected Gradient Descent (PGD) \cite{madry2017towards}}. Madry \etal (2017) proposed PGD, which is an iterative variant of FGSM. In each iteration, PGD follows the update rule 
\begin{equation}
    {x}^{t+1}=\textit{Clip}_{(x,\epsilon)} (x^t+ \alpha sign(\bigtriangledown_x L(\theta,x^t,y)).
\end{equation} The clip function $\textit{Clip}_{(x,\epsilon)}$ keeps ${x}^{(t+1)}$ within a pre-defined perturbation range. 

\section{Experiments}
\subsection{Experiment Settings}
We conducted the experiments with public Multi-spectral Semantic Segmentation dataset \cite{ha2017mfnet}. The dataset is a public dataset made up of pairs of visible RGB and 814 $\mu$m thermal infrared(IR) that captures urban scenes. The dataset consists of 1569 RGBs and thermal imaging pairs. 820 of them were collected during the daytime and the remains were collected at night. There are 9 classes including the unlabeled background class. It has $480\times640$ spatial resolution size. The training set consists of 50\% of the daytime images and 50\% of the night images. The validation set is 25\% of the daytime and 25\% of the night. The rest of the images were used for testing. 

For the base network, we used MFNet \cite{ha2017mfnet} as our baseline network. It has an encoder-decoder network. Since the dataset has a  high resolution, a mini-inception block with dilated convolution is adopted in the encoders of MFNet.

\subsection{Evaluation Metrics}
We adopt three metrics for the quantitative evaluations of models. Two of them are for the semantic segmentation performance and the rest is used to assess robustness against adversarial attacks. The first one is the accuracy for each class, which is called as \textit{recall}. The second is the intersection over union (IOU) for each class. The average value of each metric for all classes is referred to as mAcc and mIoU, respectively. They can be obtained through the formulas below:
\begin{equation}
\begin{aligned}
mAcc={1 \over N}\sum_{i=1}^{N}{TP_i \over TP_i+FN_i},
\end{aligned}
\end{equation}

\begin{equation}
\begin{aligned}
mIoU={1 \over N}\sum_{i=1}^{N}{TP_i \over TP_i+FN_i+FN_i},
\end{aligned}
\end{equation}
where \textit{N} denotes the number of classes. The original MFNet paper did not consider unlabeled class in performance measure, but we included unlabeled background class for more accurate compare. $TP_i=\sum_{k=1}^{K}P_{ii}^{K}$, $FP_i=\sum_{k=1}^{K}\sum_{j=1,j\neq i}^{N}P_{ji}^{K}$ and $FN_i=\sum_{k=1}^{K}\sum_{j=1,j\neq i}^{N}P_{ij}^{K}$ are the true positives, false positives and false negatives for each class \textit{i}, where \textit{K} is the number of frames, $P_{ii}^K$ is the number of pixels for class $i$ in the frame. $P_{ji}^K$ is the number of pixels for class $j$ that is incorrectly classified as $i$ in the frame $k$, $P_{ij}^K$ is the number of pixels for class $i$ that is incorrectly classified as class $j$ in the frame $k$. 

The other metric is the $IoU$ ratio of robust ($IoU_R$), used for evaluating the robustness of models \cite{cho2019dapas}. It is defined as follow:
\begin{equation}
\begin{aligned}
IoU_R= {mIoU_{AP} \over mIoU_{CO}},
\end{aligned}
\end{equation} 
where $mIoU_{AP}$ denotes the $mIoU$ of an adversarial image and $mIoU_{CO}$ denotes the $mIoU$ of a clean image \cite{cho2019dapas,arnab2018robustness}

\subsection{Evaluations for Adversarial Robustness}
We observed how model performances change with different types of adversarial attacks on multimodal data fusion. We conducted experiments using FGSM \cite{goodfellow2014explaining} and PGD \cite{madry2017towards} attacks (Table \hyperref[table:1]{1} and Table \hyperref[table:2]{2}). As we increase the $l_{\infty}$ norm of $\epsilon$ perturbation, both attacks have a crucial impact on the performance of multimodal data fusion models. Not only when both of inputs were attacked, but also when only one was attacked. For example, in the case of FGSM attack, the $IoU_R$ decreased from 41.9\% to 38.1\% when we attacked on the RGB sensor. When we generated adversarial perturbations on IR, the $IoU_R$ was decreased from 54.6\% to 38.9\%. On the other hand, PGD attacks on RGB sensor made the $IoU_R$ decreased from 52.8\% to 36.9\% while it descended from 55.2\% to 40.6\% for the IR case.

From the above experiments, we verified that the multimodal data fusion model is still vulnerable to the gradient based adversarial attack(Figure \hyperref[fig:1]{1}). Even if only one of them is attacked, the overall performance of the model is remarkably degraded.

\begin{table*}[ht]
\label{table:1}
	\centering
	\caption{FGSM Attack on Multimodal Data Fusion Models ($l_{\infty}$ norm of $\epsilon$ is the perturbation when the image is normalized from 0 to 255).}
	\begin{tabular}{llllllllllll}
		\hline \hline
		\multicolumn{12}{c}{{\color[HTML]{000000} \textbf{FGSM Attack}}}                                                                                                                                                                                                                                                                                                                              \\ \hline
		\multicolumn{1}{c}{{\color[HTML]{000000} }}                       & \multicolumn{4}{c}{{\color[HTML]{000000} 10}}                                                                                                      & \multicolumn{4}{c}{20}                                                        & \multicolumn{3}{c}{40}                                                      \\ \cline{2-12} 
		\multicolumn{1}{c}{\multirow{-2}{*}{{\color[HTML]{000000} $l_{\infty}$ norm of $\epsilon$}}} & \multicolumn{1}{c}{{\color[HTML]{000000} $mAcc$}} & \multicolumn{1}{c}{{\color[HTML]{000000} $mIoU$}} & \multicolumn{2}{c}{{\color[HTML]{000000} $IoU_R$}}   & \multicolumn{1}{c}{ $mAcc$} & \multicolumn{1}{c}{$mIoU$} & \multicolumn{2}{c}{$IoU_R$}   & \multicolumn{1}{c}{ $mAcc$} & \multicolumn{1}{c}{$mIoU$} & \multicolumn{1}{c}{$IoU_R$} \\ \hline
		{\color[HTML]{000000} Attack on RGB}                              & {\color[HTML]{000000} 0.196}                   & {\color[HTML]{000000} 0.158}                   & \multicolumn{2}{l}{{\color[HTML]{000000} 0.419}} & 0.174                   & 0.144                   & \multicolumn{2}{l}{0.381} & 0.155                   & 0.132                   & 0.350                    \\ \hline
		Attack on IR                                                      & 0.221                                          & 0.206                                          & \multicolumn{2}{l}{0.546}                        & 0.155                   & 0.147                   & \multicolumn{2}{l}{0.389} & 0.136                   & 0.129                   & 0.342                   \\ \hline
		Attack on RGB+IR                                                  & 0.141                                          & 0.127                                          & \multicolumn{2}{l}{0.337}                        & 0.121                   & 0.113                   & \multicolumn{2}{l}{0.300} & 0.114                   & 0.106                   & 0.281                   \\ \hline
	\end{tabular}
\end{table*}

\begin{table*}[!t]
\label{table:2}
	\centering
	\caption{PGD Attack on Multimodal Data Fusion Models ($l_{\infty}$ norm of $\epsilon$ is the perturbation when the image is normalized from 0 to 255).}
	
	\begin{tabular}{llllllllllll}
		\hline \hline
		\multicolumn{12}{c}{{\color[HTML]{000000} \textbf{PGD Attack}}}                                                                                                                                                                                                                                                                                                                              \\ \hline
		\multicolumn{1}{c}{{\color[HTML]{000000} }}                       & \multicolumn{4}{c}{{\color[HTML]{000000} 10}}                                                                                                      & \multicolumn{4}{c}{20}                                                        & \multicolumn{3}{c}{40}                                                      \\ \cline{2-12} 
		\multicolumn{1}{c}{\multirow{-2}{*}{{\color[HTML]{000000} $l_{\infty}$ norm of $\epsilon$}}} & \multicolumn{1}{c}{{\color[HTML]{000000} $mAcc$}} & \multicolumn{1}{c}{{\color[HTML]{000000} $mIoU$}} & \multicolumn{2}{c}{{\color[HTML]{000000} $IoU_R$}}   & \multicolumn{1}{c}{ $mAcc$} & \multicolumn{1}{c}{$mIoU$} & \multicolumn{2}{c}{$IoU_R$}   & \multicolumn{1}{c}{ $mAcc$} & \multicolumn{1}{c}{$mIoU$} & \multicolumn{1}{c}{$IoU_R$} \\ \hline
		{\color[HTML]{000000} Attack on RGB}                              & {\color[HTML]{000000} 0.232}                   & {\color[HTML]{000000} 0.199}                   & \multicolumn{2}{l}{{\color[HTML]{000000} 0.528}} & 0.151                   & 0.139                   & \multicolumn{2}{l}{0.369} & 0.117                   & 0.108                   & 0.286                    \\ \hline
		Attack on IR                                                      & 0.232                                          & 0.208                                          & \multicolumn{2}{l}{0.552}                        & 0.163                   & 0.153                   & \multicolumn{2}{l}{0.406} & 0.133                   & 0.125                   & 0.332                   \\ \hline
		Attack on RGB+IR                                                  & 0.151                                          & 0.132                                          & \multicolumn{2}{l}{0.350}                        & 0.112                   & 0.104                   & \multicolumn{2}{l}{0.276} & 0.111                   & 0.103                   & 0.273                   \\ \hline
	\end{tabular}
\end{table*}

\subsection{Discussion}
Multimodal data fusion models have gradually improved their predictive performance. By using multiple input data in places with adverse weather conditions or limited visibility, multimodal data fusion models have succeeded in utilizing useful information from multiple modalities. However, our observations suggest that the benefits of improving prediction performance are not necessarily applicable to the adversarial robustness of the multimodal data fusion model. Network architectures and algorithms, which have been optimized to focus on improving the prediction, were still vulnerable to normal adversarial attacks. Our explanation is in line with the statement that adversarial examples come from useful but non-robust features. To achieve a fully complementary relation network between multiple modalities, we need to seek ways to obtain not just useful but also robust features.

\section{Conclusions}
The adversarial vulnerability of multimodal data fusion deep learning models is spotlighted in this paper. Compared to the model performance, adversarial defense of multimodal data fusion models has received little attention. Our experiments show that current multimodal data fusion model does not fully utilize the complementary information of multimodality in the sense of the adversarial robustness. Our findings prompt us to recognize the adversarial vulnerability of multimodality and open a new horizon for adversarial robustness research on multimodal deep learning.

{\small
\bibliographystyle{ieee_fullname}
\bibliography{egbib}
}

\end{document}